\documentclass{article}

\usepackage{arxiv}

\usepackage[utf8]{inputenc} 
\usepackage[T1]{fontenc}    
\usepackage{hyperref}       
\usepackage{url}            
\usepackage{booktabs}       
\usepackage{amsfonts}       
\usepackage{nicefrac}       
\usepackage{microtype}      
\usepackage{cleveref}       
\usepackage{lipsum}         
\usepackage{svg}
\usepackage{graphicx}
\usepackage[backend=bibtex,style=numeric-comp,sorting=none,firstinits=true]{biblatex}
\usepackage{doi}
\usepackage{pdflscape}
\usepackage{float}
\usepackage{multirow, makecell}

\bibliography{references}

\title{Large Language Models-Enabled Digital Twins for Precision Medicine in Rare Gynecological Tumors}
\usepackage{newfloat}
\DeclareFloatingEnvironment[name={Supplementary Table},fileext=lst,listname={List of 
	Supplementary Tables}]{suptable}
\DeclareFloatingEnvironment[name={Supplementary Figure},fileext=lst,listname={List of 
	Supplementary Figure}]{supfigure}

\usepackage{authblk}

\author[1,2,3,4]{%
	Jacqueline Lammert\thanks{\texttt{Corresponding author: Jacqueline Lammert, MD, BSc
			Mail: jacqueline.lammert@tum.de}}%
}
\author[2,5]{%
	Nicole Pfarr
}
\author[6]{%
	Leonid Kuligin
}
\author[7,8]{%
	Sonja Mathes
}
\author[1,3]{%
	Tobias Dreyer
}
\author[9]{%
	Luise Modersohn
}
\author[10]{%
	Patrick Metzger
}
\author[11,12]{%
	Dyke Ferber
}
\author[11,12,13]{%
	Jakob Nikolas Kather
}
\author[14]{%
	Daniel Truhn
}
\author[15]{%
	Lisa Christine Adams
}
\author[16]{%
	Keno Kyrill Bressem
}
\author[2,17]{%
	Sebastian Lange
}
\author[2,5]{%
	Kristina Schwamborn
}
\author[9]{%
	Martin Boeker
}
\author[1]{%
	Marion Kiechle
}
\author[1,2]{%
	Ulrich A. Schatz
}
\author[1,3]{%
	Holger Bronger
}
\author[6]{%
	Maximilian Tschochohei
}
\affil[1]{Department of Gynecology and Center for Hereditary Breast and Ovarian Cancer, Technical University of Munich (TUM), School of Medicine and Health, Klinikum rechts der Isar, TUM University Hospital, Munich, Germany}
\affil[2]{Center for Personalized Medicine (ZPM), Technical University of Munich (TUM), School of Medicine and Health, Klinikum rechts der Isar, TUM University Hospital, Munich, Germany}
\affil[3]{German Cancer Consortium (DKTK), partner site Munich, a partnership between DKFZ and TUM University Hospital, Germany}
\affil[4]{EUropean Reference Network for RAre CANcers (EURACAN) Initiative, partner site Munich, Germany}
\affil[5]{Institute of Pathology, Technical University of Munich (TUM), School of Medicine and Health, Munich, Germany}
\affil[6]{Google Cloud, Munich, Germany}
\affil[7]{Department of Dermatology and Allergy Biederstein, Technical University of Munich (TUM), School of Medicine and Health, Klinikum rechts der Isar, TUM University Hospital, Munich, Germany}
\affil[8]{Institute for History, Theory and Ethics of Medicine, University of Mainz Medical Center, Mainz, Germany}
\affil[9]{Institute of Artificial Intelligence in Medicine and Healthcare, Technical University of Munich (TUM), School of Medicine and Health, Klinikum rechts der Isar, TUM University Hospital, Munich, Germany}
\affil[10]{Institute of Medical Bioinformatics and Systems Medicine, Medical Center-University of Freiburg, Faculty of Medicine, University of Freiburg, Freiburg, Germany}
\affil[11]{Else Kroener Fresenius Center for Digital Health, Technical University Dresden, Dresden, Germany}
\affil[12]{National Center for Tumor Diseases, Heidelberg University Hospital, Heidelberg, Germany; Department of Medical Oncology, Heidelberg University Hospital, Heidelberg, Germany}
\affil[13]{Department of Medicine I, University Hospital Dresden, Dresden, Germany}
\affil[14]{Department of Diagnostic and Interventional Radiology, University Hospital RWTH Aachen, Aachen, Germany}
\affil[15]{Department of Diagnostic and Interventional Radiology, Technical University of Munich, School of Medicine and Health, Klinikum rechts der Isar, TUM University Hospital, Munich, Germany}
\affil[16]{Department of Cardiovascular Radiology and Nuclear Medicine, Technical University of Munich, School of Medicine and Health, German Heart Center, TUM University Hospital, Munich, Germany}
\affil[17]{Department of Medicine II, Technical University of Munich (TUM), School of Medicine and Health, Klinikum rechts der Isar, TUM University Hospital, Munich, Germany}
\renewcommand{\shorttitle}{LLM-Enabled Digital Twins for Precision Medicine in Rare Gynecological Tumors}

\hypersetup{
pdftitle={Large Language Models-Enabled Digital Twins for Precision Medicine in Rare Gynecological Tumors},
pdfauthor={Jacqueline Lammert, Maximilian Tschochohei},
pdfkeywords={Large Language Models (LLMs), Digital Twins, Precision Oncology, Rare Gynecological Tumors (RGTs)},
}

\begin{document}
\maketitle

\begin{abstract}
	Rare gynecological tumors (RGTs) present major clinical challenges due to their low incidence and heterogeneity. The lack of clear guidelines leads to suboptimal management and poor prognosis. Molecular tumor boards accelerate access to effective therapies by tailoring treatment based on biomarkers, beyond cancer type. Unstructured data that requires manual curation hinders efficient use of biomarker profiling for therapy matching. This study explores the use of large language models (LLMs) to construct digital twins for precision medicine in RGTs.
	\par
	Our proof-of-concept digital twin system integrates clinical and biomarker data from institutional and published cases (n=21) and literature-derived data (n=655 publications with n=404,265 patients) to create tailored treatment plans for metastatic uterine carcinosarcoma, identifying options potentially missed by traditional, single-source analysis. LLM-enabled digital twins efficiently model individual patient trajectories. Shifting to a biology-based rather than organ-based tumor definition enables personalized care that could advance RGT management and thus enhance patient outcomes.
\end{abstract}

\keywords{Large Language Models (LLMs) \and Digital Twins \and Precision Oncology \and Rare Gynecological Tumors (RGTs)}

\section{Introduction}
Rare Gynecological Tumors (RGTs), comprising over 30 distinct histological subtypes, such as sex cord stromal tumors, and uterine or ovarian carcinosarcomas, account for more than 50\% of gynecologic malignancies, presenting a major clinical challenge.\supercite{Gatta2011-kb} With an incidence rate below six per 100,000 individuals, RGTs are difficult to study through large-scale randomized trials, leading to reliance on less standardized treatment approaches such as retrospective studies, case reports, and expert opinions. This lack of robust clinical guidelines has contributed to persistently poor prognosis for patients with RGTs.\supercite{Laine2021-mz}
\par
Technological advancements in cancer diagnostics have enabled the identification of biomarkers as therapeutic targets. Biomarker-guided treatments promise to accelerate the development of precision therapeutics across tumor types, reducing the relevance of organ-based classification.\supercite{noauthor_2024-fx} The prevailing organ-centric approach to clinical trial design hinders the development of effective treatments for rare cancers with shared biomarkers.\supercite{Horak2021-io} This obstacle extends beyond rare cancers: The premature dismissal of olaparib in ovarian cancer and the seven to ten year delay in addressing Programmed Cell Death Ligand 1 (PD-L1) expressing breast and gynecological cancers with PD-L1 inhibition illustrate the need for biomarker-driven stratification for cancer treatment.\supercite{Ledermann2016-pe, Andre2024-hb}
\par
Molecular tumor boards (MTBs) are essential for interpreting biomarker profile results and matching cancer patients with appropriate therapies. This includes identifying suitable investigational drugs.\supercite{Tsimberidou2023-ub} The manual interpretation of multiple, co-occurring molecular alterations requires an in-depth understanding of their functional implications and correlations with treatment sensitivity or resistance. The rapid growth of biomedical literature and the fragmented nature of data sources make manual curation a bottleneck in efficiently translating genomic data into actionable treatment strategies.\supercite{Tsimberidou2023-ub}
\par
The data produced by MTBs is often stored in unstructured formats within electronic health records (EHRs) or other repositories, hindering their reusability for similar patients. Evaluating the effectiveness of MTB-guided treatments requires extracting follow-up data from EHRs. Unstructured text within EHRs, coupled with the lack of interoperability across healthcare institutions – particularly when MTB patients receive treatment at external facilities – renders the process labor-intensive, error-prone, and time-consuming.\supercite{Botsis2023-xq} Consequently, applying MTB insights to future patients is hindered.
\par
Advances in data capture and analysis, alongside decreasing costs in genome sequencing, are paving the way for innovative tools to manage rare or refractory cancers more effectively.\supercite{Acosta2022-qj} Digital twin technology constructs virtual representations of physical entities with dynamic, bidirectional interfaces.\supercite{Kamel_Boulos2021-fn} Initially applied in industrial engineering, digital twins can also represent the human body in healthcare. By modeling physiological processes and predicting biomarker-specific responses to treatments, digital twins can address the challenges of patient variability and the limitations of traditional one-size-fits-all approaches.\supercite{Food_undated-sb} In the case of RGT, the standard carboplatin and paclitaxel regimen, followed by chemotherapy monotreatments for subsequent lines, may not be the most effective approach.\supercite{Bogani2023-qf} Digital twins could help stratify RGT patients based on their unique biomarker profiles, enabling more tailored treatments and potentially improving outcomes, even in heavily pretreated cases.
\par
Despite their potential, the adoption of digital twins in clinical practice is constrained by the challenges associated with integrating the diverse and complex data required for their development.\supercite{Katsoulakis2024-gr} Large language models (LLMs) offer potential to assist in this process by efficiently extracting and synthesizing relevant information from diverse sources.\supercite{Adams2023-mt}
\par
In this study, we demonstrate the application of an LLM-enabled workflow for constructing  digital twins for patients with RGT, specifically metastatic uterine carcinosarcoma (UCS). 
\par
The research question was inspired by a real-world UCS case presented to a major German cancer center for evaluation of third-line treatment options. According to a consensus statement by Bogani et al., third-line monotherapy in UCS typically results in a median progression-free survival (PFS) of 1.8 months and a response rate of 5.5\%, highlighting the urgent need for novel therapeutic strategies.\supercite{Bogani2023-qf} The patient presented with a proficient mismatch repair (pMMR) carcinosarcoma with intermediate Tumor Mutational Burden (TMB) and high PD-L1 expression. Although PD-L1 positivity is common in UCS\supercite{Hacking2020-jh, Jenkins2021-no} and has been suggested as an independent prognostic factor,\supercite{Kucukgoz_Gulec2020-po} it has not been validated as a target for immunotherapy.\supercite{Bogani2023-qf}
\par
Given the potential therapeutic importance of PD-L1, we investigated outcomes in similar patients. We identified cases with high PD-L1 expression, pMMR status and low to intermediate TMB from the institutional MTB database, including non-gynecological cancers, and to expand the pool of UCS cases, from the literature. The unstructured nature of EHR and academic publications posed challenges for immediate analysis. We utilized a local LLM to extract and structure data from EHRs and a cloud-based LLM for literature data. These datasets were integrated into a unified local database, forming the foundation of an RGT Digital Twin system. This system enabled the generation of virtual representations of individual patients, allowing for the simulation of personalized treatment strategies.
\par
The RGT Digital Twin system facilitated the identification of additional therapeutic options, which were subsequently evaluated by MTB members. By integrating data from institutional sources (including non-gynecological cancers) and literature sources (to expand the pool of UCS cases), this approach provided novel insights that were not apparent from either data source alone. This integration has the potential to guide more effective treatment strategies for RGT patients and supports a shift towards a biology-based rather than organ-based definition of tumors. LLM technology enabled us to streamline the extraction, structuring, and analysis of EHR and web data, making it readily accessible for MTB evaluation. This is especially valuable in resource-limited settings like MTBs, where results can occasionally arrive too late to guide timely treatment decisions.\supercite{The_Lancet_Oncology2024-ej}

\section{Methods}
\label{sec:methods}

\subsection{Study setup}
\begin{figure}[H]
	\centering
	\includegraphics[width=380pt]{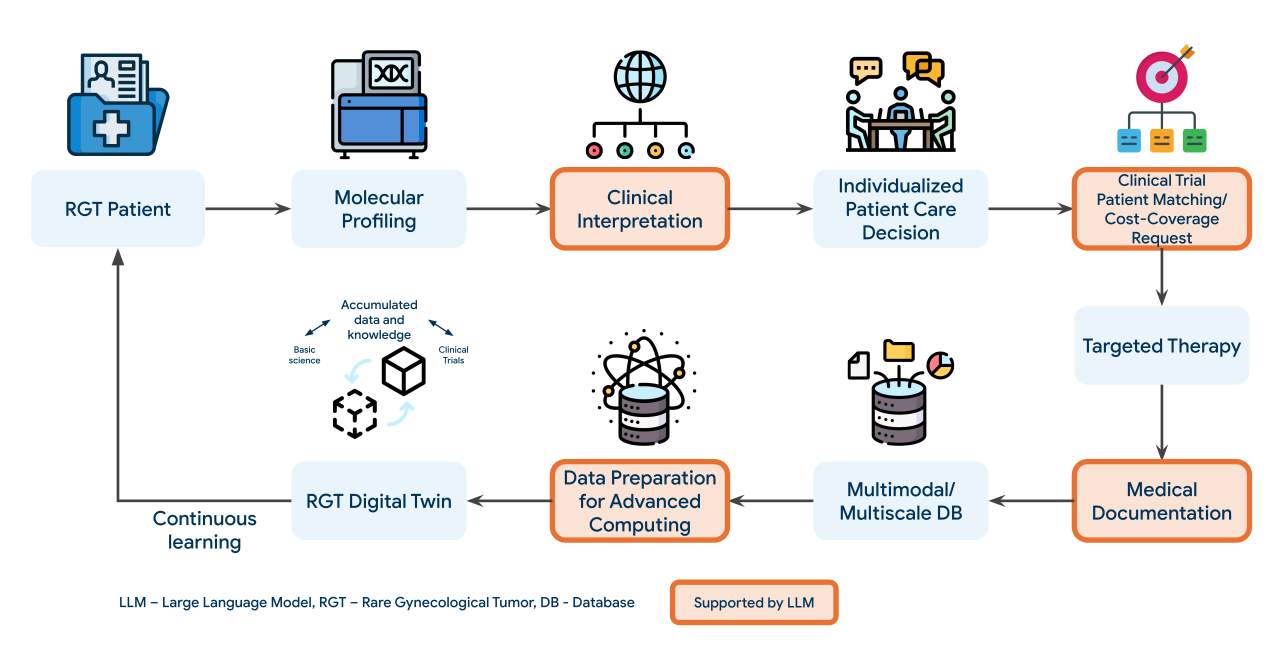}
	\caption{Workflow from RGT Patient to RGT Digital Twin}
	\label{fig:fig1}

	\raggedright 
	{\small RGT Digital Twin is a dynamic system that can integrate diverse data sources to predict individual patient trajectories. Molecular profiling identifies patient biomarkers. LLM capabilities support clinical interpretation of molecular profiles, patient matching to clinical trials, reasoning for cost-coverage requests, medical documentation, and data preparation for advanced computing.\par
	Advanced computing techniques such as classification and regression algorithms enable the creation and exploration of Digital Twin models. By adjusting parameters such as biomarker expression or previous treatment strategies, clinicians can model potential patient outcomes and determine suitable treatment strategies. The RGT Digital Twin then integrates outcome data back into the RGT Patient EHR, supporting a continuous learning process.}
\end{figure}

We employed an RGT Digital Twin system to create personalized treatment suggestions for UCS. A real-world patient case, along with molecular profiling data, was analyzed using the RGT Digital Twin system. The findings were then compared to analogous cases drawn from institutional and public databases. Treatment options were discussed at the MTB to inform individualized care decisions. Post-treatment outcomes were documented in the patient’s EHR and updated for the individual RGT Digital Twin to improve future predictions. The RGT Digital Twin provided rationale for cost coverage requests and supported study inclusion decisions. Refer to Figure \ref{fig:fig1} for an overview of the study process.

\subsection{Patient description and research question}
The patient is a 77-year-old woman with metastatic UCS, initially diagnosed with FIGO IIIC2 UCS at age 66. Six years after surgery and adjuvant chemotherapy with carboplatin and paclitaxel, the patient experienced a recurrence in the cervical lymph nodes and pelvis. A cervical lymph node biopsy confirmed the recurrence, and the patient underwent the same chemotherapy regimen followed by MTB presentation in 2021 (see Supplementary Table \ref{tab:suptable1} for detailed results). Genomic profiling was conducted using the TruSight Oncology 500 (TSO 500) and TruSight Tumor 170 (TST 170) panels. By analyzing a wide range of cancer-related genes, these panels facilitate the discovery of potential therapeutic targets. Molecular profiling revealed an intermediate TMB of 6.3 mutations/megabase, high PD-L1 expression (Combined Positive Score, CPS: 41), and a pMMR status. The patient's high PD-L1 expression prompted us to investigate the potential efficacy of PD-L1 inhibitors in metastatic tumors, regardless of primary tumor site or regional drug approval status. To this end, we searched our institutional MTB database for analogous cases, not restricted to gynecological cancers, and expanded our cohort with additional UCS cases identified in the literature.

\subsection{Data collection}
EHR data obtained from Technical University of Munich (TUM) University Hospital was the primary data source. Data downloaded from web-based repositories through institutional access extended the dataset. These repositories included \textit{PubMed}, \textit{ClinicalTrials.gov}, and clinical practice guidelines from the National Comprehensive Cancer Network (NCCN) and the German Cancer Society (DKG), which ensured adherence to up-to-date clinical standards.
\par
\begin{figure}[H]
	\centering
	\includegraphics[width=380pt]{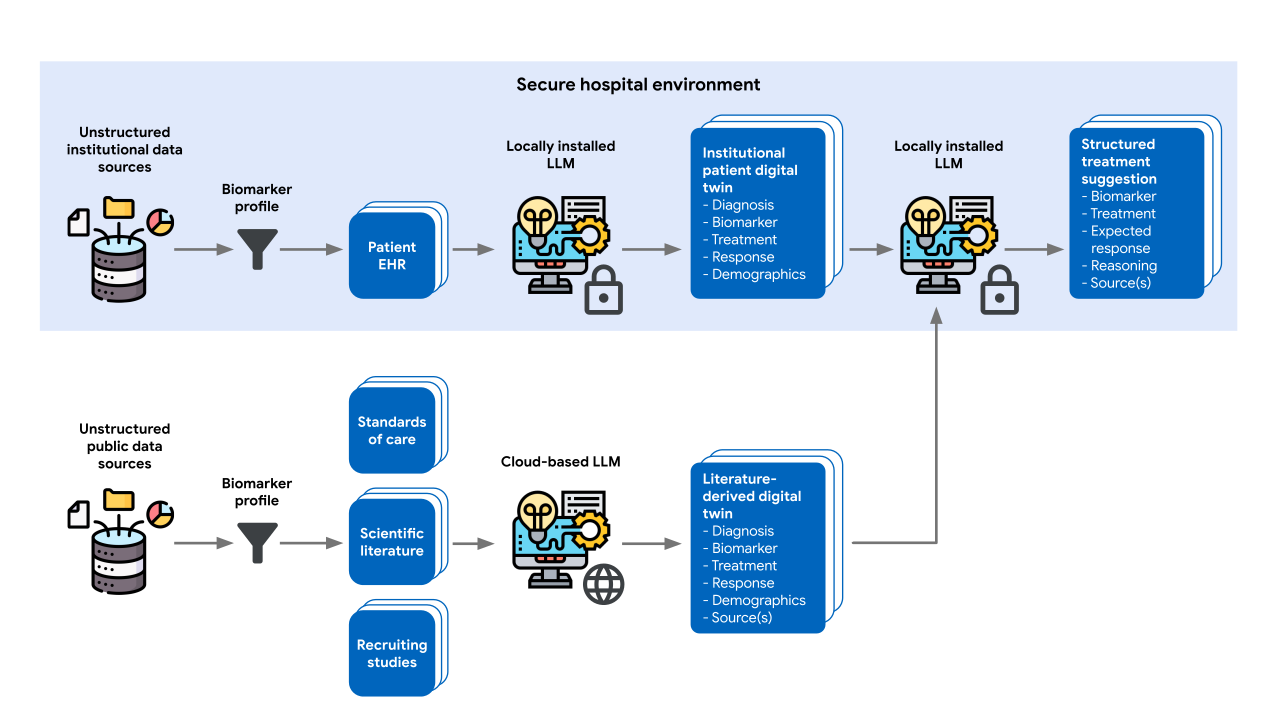}
	\caption{LLM-enabled RGT Digital Twin Pipeline}
	\label{fig:fig2}
	
	\raggedright 
	{\small To obtain institutional patient data and matching patient profiles from literature, we first filtered institutional records and public data sources (e.g., \textit{PubMed}) by biomarker profiles and primary tumor site. We then extracted structured patient data from EHR using a locally deployed, privacy-preserving LLM, and extracted similar data from published literature using a cloud-based LLM. By utilizing a broader patient population than what is available in institutional data, the RGT Digital Twin system generated personalized treatment plans for MTB evaluation. This method revealed additional treatment options that might have been missed when considering each data source alone.}
\end{figure}

A two-stage approach was employed for extracting structured, actionable data from source files. First, a locally deployed LLM system extracted relevant data from institutional EHRs. Second, a cloud-based LLM system processed documents from web-based repositories. Afterwards, the extracted structured dataset was made available to clinicians and the locally deployed LLM for further analysis. This process is shown in Figure \ref{fig:fig2}.

\subsection{Identification of analogous institutional cases}
Analysis of the institutional MTB database at the TUM University Hospital identified cases analogous to the presented UCS patient. The analysis included patients discussed at the MTB between September 2017 and July 2024. Eligibility criteria for screening included high PD-L1 expression (CPS $\ge$40) and availability of MMR and TMB status. To avoid bias, we excluded patients with high TMB and deficient Mismatch Repair, which is known to be responsive to ICI and approved as Food and Drug Administration targets for ICI therapy. Similarity to the UCS case was determined based on medical discipline (gynecological oncology), or histopathological features independent of gender or origin (carcinosarcoma or sarcomatoid carcinoma morphology). Given the shared molecular and genomic characteristics between UCS and high-grade serous ovarian and endometrial carcinomas, gynecological cancers were chosen as a criterion for similarity.\supercite{Bogani2023-qf} Due to the unclear clinical utility of distinguishing carcinosarcoma from sarcomatoid carcinoma, the institutional MTB members combined them under the category morphology. Patients were included in the final analysis if they met all of the following conditions: CPS $\ge$40, pMMR status, TMB <15 mutations/megabase, and conformance to at least one of the predefined similarity parameters.

\subsection{Institutional patient data extraction pipeline}
EHR of selected patients were processed in a secure hospital environment. Documents varied in format, from (handwritten) medical notes to obituaries. Ten attributes were extracted from documents for each patient to form the RGT Digital Twin. Supplementary Table \ref{tab:table2} shows the full list of attributes. Optical Character Recognition (OCR) was performed using Tesseract.\supercite{Hegghammer2022-dv} Raw text was processed with a locally deployed version of pre-trained LLM \textit{gemma-2-27b-it}, chosen for its ability to run locally while maintaining strong performance on medical texts.\supercite{Gemma_Team2024-iv} This privacy-preserving architecture ensured that patient data would not leave the local clinic environment. In-context learning was used to adapt the LLM to the task at hand. With this method, LLMs receive extensive instructions in their prompt, e.g., in the form of example input and output. This improves their recall and precision.\supercite{Xie2021-sl} The study was approved by the local ethics committee of TUM (Reference No. 2023-486-S-SB).

\subsection{Literature extraction pipeline}
To extend the limited sample of analogous digital twins available in institutional EHR, a systematic literature search was conducted on \textit{PubMed} using the terms ‘uterine carcinosarcoma’ and ‘endometrial carcinosarcoma’. Studies and case reports that included individual clinical follow-up data on patients with UCS treated with ICI were considered for inclusion in the analysis. Potential alternative treatment options and therapeutic targets were identified through a comprehensive review of national (DKG) and international (NCCN) oncological guidelines, \textit{PubMed}-indexed publications, as well as the \textit{ClinicalTrials.gov} database.
\par
Data points were extracted in the structure shown in Supplementary Table \ref{tab:suptable2}. The sample size was captured as an additional data point. Additionally, the LLM was instructed to extract the main treatment recommendation based on the patient profile in the paper. General purpose LLM \textit{Google-Gemini-1.5-Pro} was selected for this task due to its large context window, which enabled it to process all files in the sample without splitting them into smaller chunks.\supercite{Gemini_Team2024-pt} Since no institutional patient data was processed in this step, use of public cloud resources was permitted. The LLM was instructed to return results in the form of a JSON object. The LLM processed all documents sequentially, with each document processed in-context. Outputs were exported to a Pandas dataframe on the local machine in the secure hospital environment for convenient analysis by clinicians.

\subsection{Construction of LLM-Enabled Digital Twin System}
Next, extracted data points were stored in a database in the secure hospital environment that constituted the patient's digital twin. Clinicians were able to model potential outcomes for their patients and determine suitable treatment strategies by reviewing treatment outcomes from patients with similar biomarkers and treatment history. Additionally, they were able to employ the local LLM to combine treatment strategies identified from web sources with the patient's digital twin, creating personalized treatment recommendations. After selecting a treatment strategy, the database served as evidence for MTB evaluation, clinical trial matching, and creation of cost coverage requests with health insurance providers. 
\par
The developed pipeline is illustrated in Figure \ref{fig:fig2}.

\subsection{Analysis}
Clinical characteristics, treatment regimens, duration of therapy, treatment responses, PFS, and overall survival (OS) were systematically collected from patient records and reports. Treatment response was captured from radiology reports and categorized as complete response (CR), partial response (PR), stable disease (SD), mixed response (MR), or progressive disease (PD). For additional treatment strategies, outcomes were summarized for each therapeutic approach. Cases were sequentially numbered, starting with those retrieved from the institutional MTB database, followed by cases identified from the literature.
\par
Formal statistical analysis to evaluate the accuracy of LLM data retrieval was performed by experts. Due to the large amount of data processed by the Digital Twin pipeline, we adopted human-in-the-loop reviews, an important aspect of machine learning. \supercite{Mosqueira-Rey2023-kz} To ensure that no information was missed during extraction, a sample-based review of LLM output was performed according to machine learning leading practice. \supercite{Marshall2019-zc} For institutional data, experts reviewed all attributes extracted from EHR by the LLM for correctness. For public research data, experts reviewed a random sample of attributes extracted from scientific studies for correctness. Additionally, all attributes that were used by the LLM to construct the literature-derived digital twins were manually reviewed. Afterwards, accuracy, precision, recall, and F1 scores of LLM extraction were calculated. Finally, all treatment recommendations generated by the LLM were manually reviewed and corrected by human experts. The data extraction review panel included two bioinformaticians and two gynecological oncologists with five and 16 years of clinical experience.
\par
A panel of five MTB members, including three clinicians, one pathologist, and one biologist, along with a senior gynecological oncologist, evaluated the personalized treatment recommendations generated by the RGT Digital Twin system.
\par
All statistical analyses were conducted using Pandas and SciPy libraries in Python (Version 3.10.12). The full code and documentation is available on \href{https://github.com/LammertJ/RGT-Digital-Twin/}{GitHub}.

\section{Results}
\label{sec:results}

\subsection{Patient cohort}
A retrospective analysis of 1821 cases discussed at the institutional MTB between September 2017 and July 2024 was conducted. Among these, 132 cases exhibited high PD-L1 expression (CPS $\ge$40), encompassing 28 different tumor entities. The analysis was restricted to patients with TMB <15 mutations/megabase and pMMR status with either gynecological cancers or carcinosarcoma/sarcomatoid carcinoma, resulting in a cohort of nine patients. Of these, seven patients received ICI therapy and were included in the study. The cohort comprised six females and one male aged 32 to 83 years at MTB presentation. Given that the similarity analysis focused on biomarker profiles and cancer morphology, the male patient's inclusion was appropriate. His sarcomatoid carcinoma aligned with the other inclusion criteria, regardless of his gender or cancer type.

\subsection{Data extraction}
89 EHR documents were extracted for the patient cohort (median: 11, range: 9-21). Documents had a median of two pages and 4,340 characters. Documents contained 70 data points for the selected attributes. Experts reviewed all extractions in the sample. The local LLM achieved accuracy of 0.76, precision of 0.96, recall of 0.78 and F1 of 0.86. The highest accuracy was achieved for 'diagnosis' and 'ICI treatment' (1.00). Low recall occurred in 'previous treatment' (0.29) and 'PFS' (0.14), mainly due to parsing errors in order and dates of previous treatments. See Table \ref{tab:table2} for full results of the analysis.
\par 
Document analysis revealed that primary tumor sites included metastatic UCS (n=1), metastatic cervical cancer (n=4; three squamous cell carcinoma, one adenocarcinoma), metastatic uterine serous carcinoma (n=1), and metastatic, undifferentiated sarcomatoid carcinoma of the pancreas (n=1). Patients exhibited high PD-L1 expression with a median CPS of 75 (range: 40-95) and a median TMB of 5.5 (range: 0-11). Median follow-up duration was 48 months (range: 15-132 months). Detailed baseline clinical characteristics are presented in Supplementary Table \ref{tab:suptable3}.
\par
The LLM-based systematic literature research yielded a dataset of 663 scientific documents. Files had a median of seven pages and 27,995 characters, with a maximum of 934,513 characters. The LLM extracted 7,956 attributes from scientific documents. Attribute extraction was reviewed with a random sample of n = 352 (Z = 1.96, N = 7,956, e = 0.05, P = 0.5). The cloud-based LLM achieved accuracy of 0.98, precision of 1.00, recall of 0.97, and F1 of 0.98. Lowest recall was observed in 'PFS' (0.77).
\par
The LLM system identified 15 studies reporting ICI treatment in UCS, encompassing a total of 215 patients. While seven of the studies did not exclusively enroll UCS patients, four provided stratified outcomes for UCS cases. Phase II studies that provided stratified analysis for UCS patients showed objective response rates between zero and ten percent for ICI treatment. None of the seven studies allowed for individual patient-level data extraction to create digital twins (see details on these seven studies in Supplementary Table \ref{tab:suptable4}).
\par
PD-L1 status was reported in 10 of the 215 literature-derived UCS cases treated with immunotherapy, with three cases exhibiting PD-L1 positivity. This limited sample size precluded stratified analysis. Notably, two of the PD-L1-positive UCS patients harbored dMMR and one had a high TMB, both of which are known to influence ICI treatment response.
\par
Further analysis of the 15 identified studies yielded eight studies with individual patient follow-up data, comprising a total of 14 cases. The median age of these literature-derived patients was 63 years (range: 55-68 years).

\subsection{Treatment response outcomes for 21 individual patients}
In the institutional cohort, seven patients received ICI therapy: five with pembrolizumab monotherapy, one with pembrolizumab plus lenvatinib, and one with ipilimumab plus nivolumab. ICI therapy was initiated on average in the third line (range: 2-4). Median PFS was 6 months (range: 1-48). One patient remained disease-free after 45 months, two continued to respond, one received a subsequent  treatment line, and three had died.
\par
Treatments in the 14 cases of the literature-derived cohort consisted of pembrolizumab (n=4), pembrolizumab plus lenvatinib (n=7), pembrolizumab plus lenvatinib plus letrozole (n=1), PD-1/Cytotoxic T-lymphocyte associated protein 4 inhibitors (n=1), and avelumab plus axitinib (n=1). ICI was typically given in the third line (range: 2-5). Median PFS was 4 months (range: 0.9-15), and median OS was 9.9 months (range: 2.1-48). At data cut-off, six patients were alive, seven had died, and one had unknown status.
\par
Table \ref{tab:table2} provides a summary of ICI treatment response outcomes for all 21 cases.

\subsection{RGT Digital Twins enable predictive modeling of individualized patient treatment strategies}
To inform personalized treatment planning for the UCS patient (case 1), digital twins were created based on 21 evaluable patients. Treatment outcomes were  predicted based on a database of additional 404,265 cases derived from scientific papers (n = 655). Potential treatment strategies were predicted for a patient with UCS with disease progression following third-line pembrolizumab monotherapy.
\par
Supplementary Figure \ref{fig:supfig1} presents treatment-relevant biomarkers after progression on standard-of-care combination treatment with carboplatin and paclitaxel identified by the digital twin system.
\par
The digital twin system tailored treatment recommendations based on the patient's specific tumor characteristics, treatment history, and geographic location. Considering the patient's ongoing pembrolizumab therapy, the system suggested testing for Folate Receptor Alpha (FR$\alpha$) to assess potential eligibility for an off-label treatment regimen currently under clinical investigation. This trial investigated the combination of mirvetuximab soravtansine and pembrolizumab in FR$\alpha$-positive UCS, with eligibility criteria including pMMR status and prior pembrolizumab progression. However, the trial was no longer recruiting participants and was limited to the United States.\supercite{noauthor_undated-do} Therefore, the digital twin system suggested considering off-label use of this regimen for the patient. A previous evaluation (2021) identified HER2 amplification in the patient's tumor, a biomarker linked to high objective response rates to trastuzumab deruxtecan.\supercite{Nishikawa2023-hj} Due to the potential for HER2 status to evolve, the system recommended confirming this finding through a new biopsy. \supercite{Yoshida2023-bl} Additionally, the digital twin system suggested evaluating Melanoma-Associated Antigen A4 (MAGE-A4) and Preferentially Expressed Antigen in Melanoma (PRAME), biomarkers frequently expressed in UCS.\supercite{Resnick2002-mj, Alrohaibani2024-ia} Ongoing research explores targeted therapies for these markers. Three relevant clinical trials were accessible within the patient’s geographic area. To monitor disease progression, the system recommended continued tracking of serum Cancer Antigen-125 (CA-125) levels based on its established correlation with disease progression identified in the patient's 2021 EHR data.\supercite{Huang2007-aj}
\par
Potential treatment trajectories for treatment line four derived from the Digital Twin pipeline are demonstrated in Table \ref{tab:table3}.

\section{Discussion}
\label{sec:discussion}
Extracting meaningful data from unstructured medical text is a prerequisite for precision medicine. In this study, we implemented an LLM-based extraction pipeline to systematically retrieve, structure, and analyze data from real-world EHRs and online sources to support and evaluate diagnostic and targeted therapeutic strategies for constructing patient-specific digital twins for metastatic UCS.
\par
The LLM-based extraction pipeline facilitated timely and accurate synthesis of all relevant full-text scientific publications available through institutional access up to August 15, 2024. The cloud-based LLM achieved accuracy of 0.98 on a complex corpus of medical literature, close to the 0.96 observed by other researchers.\supercite{Konet2024-jg} This enabled the generation of evidence-based recommendations and predictive insights grounded in the latest research. Key gaps were observed in extraction of complex data structures, with recall of 'PFS' (0.77) and 'OS' (0.95) below the overall recall of 0.97. This was due to the fragmented and unstructured way of reporting PFS and OS. Sentences such as "Patient survived for 14 months with the residual tumor post-relapse,"\supercite{Yano2020-vq} make it challenging to accurately determine PFS, as it requires estimation based on prior treatments and the number of treatment cycles. However, estimated PFS may not be accurate if treatment cycles were prolonged. This highlights the challenge of extracting precise outcome data when the primary source lacks comprehensive reporting. We strongly advocate for standards in reporting treatment outcomes, e.g., by clearly stating PFS in months and not date ranges.
\par
In institutional data, unstructured EHR impeded the extraction of key clinical information. This limitation delayed the integration of institutional patient data for informing the management of similar cases. Phase II trials neglecting biomarker-stratification in patients with UCS yielded low objective response rates to ICI therapy, ranging from only zero to ten percent.\supercite{Rubinstein2023-we,Lheureux2022-bf} For our UCS patient, this bottleneck might have precluded ICIs based on high PD-L1 expression, despite the fact that pembrolizumab proved highly efficacious with no adverse effects in this patient. The local LLM system was able to extract structured follow-up data from EHRs across a diverse and complex set of medical documents. While it achieved lower accuracy than the cloud-based model at 0.76, this is in line with the performance of similar models on complex EHR.\supercite{Saab2024-di} Notably, recall was high across most attributes, with the most critical gap noted in 'biomarkers' at 0.57. The LLM achieved full recall for all biomarkers given as examples for in-context learning, but did not recognize biomarkers that were not explicitly mentioned (e.g., \textit{BRAF} for case four). The LLM again achieved lowest recall for 'PFS' at 0.14. This is due to the highly unstructured and fragmented way of reporting PFS, often across multiple documents.
\par
The European Society for Medical Oncology Precision Medicine Working Group recently established criteria for evaluating the tumor-agnostic potential of molecularly guided therapies, mandating an ORR of $\ge$20\% in at least one of five patients across at least four investigated tumor types, with a minimum of five evaluable patients per type.\supercite{noauthor_2024-fx} Our institutional MTB database identified six analogous cases involving four additional tumor types, most of which exhibited durable responses to ICIs. While the limited number of evaluable patients per tumor type in our single-institution cohort restricted the statistical power, an LLM-driven literature review highlighted an underreporting of PD-L1 expression in UCS in studies conducted to date, despite the known high prevalence of PD-L1 positivity in this malignancy.\supercite{Hacking2020-jh, Jenkins2021-no} This underreporting impeded our ability to assess the predictive value of PD-L1 to guide ICI treatment in UCS. Despite the limited sample size of our institutional cohort, the promising outcomes observed suggest that targeting PD-L1 expression in RGT may be a viable therapeutic strategy. The inclusion of diverse tumor types in our institutional cohort further strengthens the role of PD-L1 inhibition in both gynecological and non-gynecological cancers, making it a potential tumor-agnostic marker. Combining PD-1 blockade with bispecific antibodies could offer a promising approach for treating tumors that have not responded to checkpoint inhibitor monotherapy.\supercite{Merz2024comeback}
\par
To inform treatment strategies in the event of disease progression, we constructed 21 individualized digital twins, including 7 from our institutional database and 14 from the literature, and queried an LLM-derived database containing 404,265 patient cases. Although our systematic \textit{PubMed} search was specifically limited to the terms "uterine carcinosarcoma" and "endometrial carcinosarcoma," the resulting sample also included other uterine and ovarian malignancies. This is because UCS is frequently reported within the broader context of clinical trials involving more common gynecologic cancers.
\par
The RGT digital twin system generated individualized trajectory predictions for various targeted therapies within a secure local environment that respects patient data privacy, offering guidance on further diagnostics, potential treatment options and continued treatment monitoring with serum CA-125. Additionally, since our real-world cohort comprised only White patients, being able to extract data on patients of other races from a vast corpus of literature helped us validate the generalizability of our treatment recommendations.\supercite{Arora2022-gq}
\par
This study successfully demonstrated the utility of RGT digital twins for individualized treatment prediction and response modeling. The digital twin not only provided generalized recommendations for additional diagnostic testing but also incorporated specific clinical details from the patient’s treatment history – such as prior pembrolizumab administration – to refine eligibility assessments for targeted therapies. The extraction and analysis of follow-up data revealed that, following the MTB recommendation, the patient received pembrolizumab due to high PD-L1 expression and exhibited a sustained partial response for over 30 months.
\par
This study had several limitations. Firstly, despite a large dataset, the combination of stringent similarity criteria, a limited institutional cohort, and underreported PD-L1 status in published UCS cases prevented us from stratifying patients by PD-L1 status. Efficacy of ICI treatment in PD-L1–positive UCS remains uncertain, and current trials lack PD-L1 as a stratification factor. Bogani et al. listed nine clinical trials currently exploring ICI treatment in UCS, many of which are nearing completion.\supercite{Bogani2023-qf} None of these trials included PD-L1 expression as a stratification factor. Our findings could inform the design of future trials that specifically evaluate ICI efficacy in pMMR UCS with high PD-L1 expression and low to intermediate TMB. Secondly, only somatic biomarkers were included, potentially underestimating clinical actionability by excluding germline mutations, such as \textit{BRCA1/2}, which are predictive of PARP inhibitor response.\supercite{Mendes_Gomes2024-ic} Thirdly, a local LLM was used for data extraction, impacting extraction performance due to its smaller size and no fine-tuning on German medical texts.\supercite{Han2023-xn} Lack of a German-language equivalent to the English-language MIMIC labeled medical record dataset\supercite{Johnson2023-dx} precluded fine-tuning our own model. Finally, the German Network for Personalized Medicine (DNPM) data model is under revision,\supercite{Siebrasse_undated-dt} necessitating the use of a custom data model for this study and highlighting the importance of future validation for compatibility with DNPM v2.
\par
National and international collaborative initiatives, such as the DNPM Data Integration Platform\supercite{Illert2023-gy} and the Molecular Tumor Board Portal by Cancer Core Europe,\supercite{Tamborero2022-xj} aim to enhance MTB decision-making by standardizing and harmonizing data collection across institutions. These platforms stand to benefit substantially from the integration of LLM-based extraction pipelines, which could facilitate the automated extraction of both baseline and follow-up data, thereby enabling the real-time utilization of MTB data across different healthcare systems. Once the DNPM database is fully operational, clinical narratives derived from EHR data could be transformed into HL7 Fast Healthcare Interoperability Resources (FHIR), streamlining interoperability and reducing the biases and costs associated with manual documentation.\supercite{Li_Yikuan2024-eb} Such automation would enable the analysis of larger patient cohorts, thereby providing the statistical power necessary for accurate treatment predictions in rare cancers, a critical step towards advancing personalized oncology. The outcome data could subsequently be used to inform both preclinical research and stratified clinical trials.
\par
Our LLM-enabled precision oncology approach can inform more effective treatment strategies for RGT patients and supports a paradigm shift from organ-based to biology-based tumor classification.
\par
Given the increasing volume and complexity of precision oncology data from MTBs, and the limited availability of precision oncologists to translate this abundance of information into clinically meaningful actions,\supercite{Tsimberidou2023-ub} there is an urgent need for advanced digital tools to facilitate the extraction, structuring, and analysis of large datasets.\supercite{The_Lancet_Oncology2024-ej} Our proof-of-concept study demonstrates the potential of LLMs to efficiently synthesize relevant information for MTB evaluation.
\par
While this study focused on RGTs, the LLM-enabled digital twin approach holds potential for a wide range of refractory cancers. By accurately predicting individual patient trajectories, these digital twins can inform personalized diagnostics and treatment strategies in a timely and cost-effective manner, potentially improving patient outcomes.
\section*{Additional information}
\subsection*{Ethics statement}
Patients from the institutional MTB were included in a clinical registry that allowed for retrospective analyses of clinical and molecular anonymized data in accordance with the Declaration of Helsinki. The retrospective analysis was approved by the Ethics commission of the Medical Faculty of the Technical University of Munich (Reference No. 2023-486-S-SB).
\par
All web-based research procedures were conducted exclusively on publicly accessible, anonymized patient data and in accordance with the Declaration of Helsinki, maintaining all relevant ethical standards.
\subsection*{Acknowledgements}
JL is a fellow of the TUM School of Medicine Clinician Scientist Program (project no. H-08). JL acknowledges a 5,000 USD funding in Google Cloud credits from Google's Gemma Academic Research Program. Figures 1 and 2 incorporate icons obtained from Flaticon.com.
\subsection*{Author contributions}
Conceptualisation: JL, MT; project administration: JL, MT; resources: JL, NP, KS and SL; validation: JL, NP, LK, KS, SL, US, HB, MT; writing – original draft preparation: JL, MT; writing –review \& editing: SM, TD, LM, PM, DF, JNK, DT, LCA, KKB, MB, MK, US, HB. All authors contributed scientific advice and approved the final version of the manuscript.
\subsection*{Funding}
JL is supported by the TUM School of Medicine and Health Clinician Scientist Program (project no. H-08). JL receives support through the DKTK School of Oncology Fellowship. PM is funded by the German Federal Ministry of Education and Research through the funding programs Medical Informatics Initiative and National Decade against Cancer (PM4Onco; 01ZZ2322A). JNK is supported by the German Federal Ministry of Health (DEEP LIVER, ZMVI1-2520DAT111; SWAG, 01KD2215B), the Max-Eder-Programme of the German Cancer Aid (grant \#70113864), the German Federal Ministry of Education and Research (PEARL, 01KD2104C; CAMINO, 01EO2101; SWAG, 01KD2215A; TRANSFORM LIVER, 031L0312A; TANGERINE, 01KT2302 through ERA-NET Transcan), the German Academic Exchange Service (SECAI, 57616814), the German Federal Joint Committee (Transplant.KI, 01VSF21048) the European Union’s Horizon Europe and innovation programme (ODELIA, 101057091; GENIAL, 101096312) and the National Institute for Health and Care Research (NIHR, NIHR213331) Leeds Biomedical Research Centre. DT is funded by the German Federal Ministry of Education and Research (TRANSFORM LIVER, 031L0312A), the European Union’s Horizon Europe and innovation programme (ODELIA, 101057091), and the German Federal Ministry of Health (SWAG, 01KD2215B). KKB is supported by the European Union’s Horizon Europe and innovation programme (COMFORT, 101079894), Bayern Innovative and Wilhelm-Sander Foundation. MK reports research grants from Sphingotec, Deutsche Krebshilfe, DFG, Senator Roesner Foundation, Dr. Pommer-Jung Foundation, Waltraut Bergmann Foundation, Bavarian State Ministry of Economy, BMBF, G-BA Innovation Fonds. HB reports grants from Deutsche Forschungsgemeinschaft (DFG, German Research Foundation) during the conduct of the study. No other funding is disclosed by any of the authors. 
\subsection*{Competing interests}
JL received honoraria by the Forum for Continuing Medical Education (FomF) and Novartis for delivering educational lectures. NP has received honoraria for consulting and advisory board participation by AstraZeneca, MSD, GSK, BMS, LabCorp, Illumina, Janssen Cilag, QuIP, and Novartis. DF has received a research grant from OpenAI and holds shares and is an employee at Synagen GmbH. JNK declares consulting services for Owkin, France; DoMore Diagnostics, Norway; Panakeia, UK, and Scailyte, Basel, Switzerland; furthermore JNK holds shares in Kather Consulting, Dresden, Germany; StratifAI GmbH, Dresden, Germany, and Synagen GmbH, Germany, and has received honoraria for lectures and advisory board participation by AstraZeneca, Bayer, Eisai, MSD, BMS, Roche, Pfizer and Fresenius. DT received honoraria for lectures by Bayer and holds shares in StratifAI GmbH, Germany and Synagen GmbH, Germany. KKB has received honoraria for lectures by GE HealthCare and Canon Medical Systems Corporation and serves as advisor for the EU Horizon 2020 project LifeChamps (875329) and the EU IHI Project IMAGIO (101112053). SL has received honoraria for lectures and advisory boards from Taiho Oncology, AstraZeneca, MSD and Janssen-Cilag and provides medical consulting services for NEED Inc. MK has received fees from Springer Press, Biermann Press, Celgene, AstraZeneca, Myriad Genetics, TEVA, Eli Lilly, GSK, Seagen, AllergoSan, FomF, Roche, BESINS, Bayer AG. KS has received honoraria for lectures and advisory boards from BMS, MSD, AstraZeneca, Janssen-Cilag, Roche and Solution akademie GmbH. MK has received honoraria for consulting and advisory board participation by Myriad Genetics, Bavarian KVB, DKMS Life, BLAEK, TEVA, Exeltis, Roche, BESINS, Bayer AG. MK holds shares in AIM GmbH, in-manas GmbH, Therawis Diagnostic GmbH. HB has received personal fees from Roche, AstraZeneca, Gilead, and GlaxoSmithKline outside the submitted work. The authors have no additional financial or non-financial conflicts of interest to disclose.
\subsection*{Data availability statement}
Institutional data is available upon reasonable request. All notebooks and prompts used in this study are publicly available at \href{https://github.com/LammertJ/RGT-Digital-Twin/}{GitHub}.
\newpage
\section*{Figures}

\newpage
\section*{Tables}

\begin{table}[h!]
	\caption{Evaluation of LLM performance for EHR and literature record extraction}
	\centering
		\scalebox{.6}{
	\begin{tabular}{|l|l|l|l|l|l|l|l|l|l|l|}
		\toprule
		Source File & Data point & Observations & True Positive & True Negative & False Positive & False Negative & Accuracy & Precision & Recall & F1 \\
		\midrule
		EHR & Age & 7 & 6 & 0 & 1 & 0 & 0.86 & 0.86 & 1.00 & 0.86 \\
		EHR & Gender & 7 & 7 & 0 & 0 & 0 & 1.00 & 1.00 & 1.00 & 1.00 \\
		EHR & Race & 7 & 7 & 0 & 0 & 0 & 1.00 & 1.00 & 1.00 & 1.00 \\
		EHR & Diagnosis & 7 & 7 & 0 & 0 & 0 & 1.00 & 1.00 & 1.00 & 1.00 \\
		EHR & Biomarkers & 7 & 4 & 0 & 0 & 3 & 0.57 & 1.00 & 0.57 & 0.73 \\
		EHR & Previous treatments & 7 & 2 & 0 & 0 & 5 & 0.29 & 1.00 & 1.00 & 1.00 \\
		EHR & Study treatments & 7 & 7 & 0 & 0 & 0 & 1.00 & 1.00 & 1.00 & 1.00 \\
		EHR & Study treatment response & 7 & 5 & 0 & 1 & 1 & 0.71 & 0.83 & 1.00 & 0.91 \\
		EHR & PFS {[}months{]} & 7 & 1 & 0 & 0 & 6 & 0.14 & 1.00 & 0.14 & 0.25 \\
		EHR & OS {[}months{]} & 7 & 7 & 0 & 0 & 0 & 1.00 & 1.00 & 1.00 & 1.00 \\
		EHR & TOTAL & 70 & 53 & 0 & 2 & 15 & 0.76 & 0.96 & 0.85 & 0.91 \\
		\midrule
		Source File & Data point & Observations & True Positive & True Negatives & False Positive & False Negatives & Accuracy & Precision & Recall & F1 \\
		\midrule
		Literature & Sample size & 32 & 29 & 3 & 0 & 0.00 & 1.00 & 1.00 & 1.00 & 1.00 \\
		Literature & Age & 32 & 29 & 3 & 0 & 0.00 & 1.00 & 1.00 & 1.00 & 1.00 \\
		Literature & Gender & 32 & 7 & 25 & 0 & 0.00 & 1.00 & 1.00 & 1.00 & 1.00 \\
		Literature & Race & 32 & 7 & 25 & 0 & 0.00 & 1.00 & 1.00 & 1.00 & 1.00 \\
		Literature & Diagnosis & 32 & 30 & 1 & 0 & 1.00 & 0.97 & 1.00 & 0.97 & 0.98 \\
		Literature & Biomarkers & 32 & 17 & 15 & 0 & 0.00 & 1.00 & 1.00 & 1.00 & 1.00 \\
		Literature & Previous treatments & 32 & 23 & 9 & 0 & 0.00 & 1.00 & 1.00 & 1.00 & 1.00 \\
		Literature & Study treatments & 32 & 29 & 2 & 0 & 1.00 & 0.97 & 1.00 & 0.97 & 0.98 \\
		Literature & Study treatment response & 32 & 26 & 5 & 0 & 1.00 & 0.97 & 1.00 & 0.96 & 0.98 \\
		Literature & PFS {[}months{]} & 32 & 10 & 19 & 0 & 3.00 & 0.91 & 1.00 & 0.77 & 0.87 \\
		Literature & OS {[}months{]} & 32 & 18 & 13 & 0 & 1.00 & 0.97 & 1.00 & 0.95 & 0.97 \\
		Literature & TOTAL & 352 & 225 & 120 & 0 & 7.00 & 0.98 & 1.00 & 0.97 & 0.98 \\
		\bottomrule
	\end{tabular}
	} 
	\label{tab:table1}
\end{table}

\begin{landscape}
\begin{table}[H]
	\caption{ICI treatment outcomes in 7 institutional cases and 14 literature-derived cases}
	\centering
	\scalebox{.6}{
	\begin{tabular}{|l|l|p{2.5cm}|l|l|p{2.5cm}|p{2cm}|p{2cm}|p{4cm}|p{2cm}|p{3cm}|l|l|p{2.5cm}|}
		\toprule
		ID & Reference & Diagnosis & Age\textsuperscript{a} & Race & PD-L1 Status & TMB (Mut/Mb)\textsuperscript{b} & MMR & Addtl. relevant biomarkers (All. frequency) & Treatment Line & ICI treatment (mono/combination) & Response & PFS {[}months{]} & OS {[}months{]} \\
		\midrule
		1 & Institutional & UCS & 77 & White & CPS: 41, TPS: 3\%, IC: 40\% & 6.3 & pMMR (3.6\%) & FR$\alpha$ 0.8, PR 0.03, HER2-positive & 3 & Radiotherapy + pembrolizumab (off-label) & PR & \textgreater{}30 (ongoing) & \textgreater{}132 (ongoing) \\
		2 & Institutional & CESC & 37 & White & CPS: 75, TPS: 70\%, IC: 5\% & 0 & pMMR (1.11\%) & None & 3 & Pembrolizumab (off-label) & PR & \textgreater{}49 (ongoing) & \textgreater{}79 (ongoing) \\
		3 & Institutional & CESC & 32 & White & CPS: 40, TPS: 40\%, IC : \textless{}1\% & 3.1 & pMMR (0\%) & \textit{PIK3CA} (p.E545K, 0.26), \textit{CHEK2} (p.T367Mfs*15, 0.79) & 2 & Pembrolizumab (off-label) & PD & 1 & 15 (deceased) \\
		4 & Institutional & CESC & 85 & White & CPS: 81, TPS: 80\%, IC: 1\% & 11 & pMMR (4.6\%) & \textit{BRAF} (p.D594N, 0.27), \textit{KMT2C} (p.Q192Tfs*28, 0.29) & 4 & Pembrolizumab (off-label) & PR, PD & 18 & 72 (deceased) \\
		5 & Institutional & CEAD & 37 & White & CPS: 95, TPS: 90\%, IC: 5\% & 5.5 & pMMR (4.6\%) & None & 2 & Ipilimumab/ nivolumab, nivolumab maintenance (off-label) & CR & \textgreater{}45 & \textgreater{}45 (ongoing) \\
		6 & Institutional & USC & 61 & White & CPS: 40, TPS: 30\%, IC: 8\% & 13.4 & pMMR (1.89\%) & \textit{PIK3CA} (p.E545K, 0.06), \textit{PTEN} (p.K128Rfs*6, 0.13), \textit{PTEN} (p.Y240delins*, 0.06), FR$\alpha$: 0\%, HER2: Score 0, Trop2: 100\% & 3 & Pembrolizumab + lenvatinib (in-label) & PD & 3 & \textgreater{}69 (ongoing) \\
		7 & Institutional & Undifferentiated Sarcomatoid Carcinoma of the Pancreas & 60 & White & CPS: 85, TPS: 80\%, IC: 4\% & 3.2 & pMMR (2.61\%) & \textit{KRAS} (p.G12C; 0.38) & 3 & Pembrolizumab (off-label) & PR, PD & 6 & 19 (deceased) \\
		8 & PMID: 32620662 & UCS & 65 & Asian (Japanese) & positive & n/a & dMMR/MSI-H & None & 2 & Radiotherapy + pembrolizumab & CR, PD & 10 & 16 (deceased) \\
		9 & PMID: 34401435 & UCS & n/a\textsuperscript{c} & n/a\textsuperscript{c} & negative & n/a & pMMR &  & 3 & Pembrolizumab + lenvatinib & PD & 3.3 & 9.9 (deceased) \\
		10 & PMID: 34401435 & UCS & n/a\textsuperscript{c} & n/a\textsuperscript{c} & negative & n/a & pMMR &  & 3 & Pembrolizumab + lenvatinib & PD & 0.9 & 2.8 (deceased) \\
		11 & PMID: 34401435 & UCS & n/a\textsuperscript{c} & n/a\textsuperscript{c} & positive & n/a & dMMR/MSI-H &  & 3 & Pembrolizumab + lenvatinib & PD & 1.6 & 2.4 (deceased) \\
		12 & PMID: 34401435 & UCS & n/a\textsuperscript{c} & n/a\textsuperscript{c} & negative & n/a & pMMR &  & 3 & Pembrolizumab + lenvatinib & PD & 2.6 & 2.8 (deceased) \\
		13 & PMID: 34401435 & UCS & n/a\textsuperscript{c} & n/a\textsuperscript{c} & negative & n/a & pMMR &  & 5 & Pembrolizumab + lenvatinib & PD & 1.9 & 2.1 (deceased) \\
		14 & PMID: 34401435 & UCS & n/a\textsuperscript{c} & n/a\textsuperscript{c} & negative & n/a & pMMR &  & 4 & Pembrolizumab + lenvatinib & SD & - (ongoing) & 4.4 (alive at data cut-off) \\
		15 & PMID: 34401435 & UCS & n/a\textsuperscript{c} & n/a\textsuperscript{c} & negative & n/a & pMMR &  & 3 & Pembrolizumab + lenvatinib & SD, PD & 11.2 & 12.6 (alive at data cut-off) \\
		16 & PMID: 29386312 & UCS & 55 & n/a & 1+, low positive & 169 & pMMR & \textit{POLE}-mutated & 4 & Pembrolizumab & PR & \textgreater{}12 (ongoing) & 39 (alive at data cut-off) \\
		17 & PMID: 38881561 & UCS & 68 & n/a & n/a & 6 & pMMR & \textit{PTEN} K128T, \textit{ESR1}-amplified (8/8 exons, est. 11 copies), ER positive & 2 & Pembrolizumab + lenvatinib + letrozole & PR & \textgreater{}36 (ongoing) & 45 (alive at data cut-off) \\
		18 & PMID: 30442730 & UCS & 59 & n/a & n/a & n/a & n/a &  & 2 & Pembrolizumab & MR & 4 & n/a \\
		19 & PMID: 33004543 & UCS & 66 & Asian (Japanese) & n/a & n/a & dMMR & Highly predisposing HLA haplotype for narcolepsy & 3 & Pembrolizumab & PD & 2 & Deceased, 72 days post pembrolizumab, OS n/a \\
		20 & PMID: 31149529 & UCS & 68 & n/a & n/a & n/a & n/a &  & 2 & PD-1 antibody + CTLA-4 antibody & PR & \textgreater{}5 (ongoing) & N/a, alive \\
		21 & PMID: 35434237 & UCS & 62 & n/a & n/a & 14 & pMMR & Germline \textit{NBN} mutation, (c.2117C\textgreater{}G, p.Ser706Ter) HER2-low (Score 1+) & 3 & Avelumab + axetinib & PR & \textgreater{}15 (ongoing) & 48 (alive at data cut-off) \\
		\bottomrule
	\end{tabular}
	} 
	\label{tab:table2}
	\raggedright
	\vspace{1\baselineskip}\par
	{\small 
	a. Current age at data cut-off (publication)\linebreak
	b. TMB: <5: low, 5-15: intermediate, $\ge$15: high\linebreak
	c. Patients 9-15: Median age: 63 (range: 58-64), White: 3, Black: 4. Individual data for age \& race not reported.\linebreak
	PMID: PubMed-ID\linebreak
	Note: 'n/a' entries indicate data not available for the specific case.}
\end{table}
\end{landscape}

\begin{table}[H]
	\caption{Digital twin pipeline provided the following individualized treatment targets for case 1}
	\centering
	\scalebox{.6}{
	\begin{tabular}{|l|p{3.5cm}|p{3cm}|p{3cm}|p{7cm}|p{4cm}|}
		\toprule
		Biomarker & Prevalence in UCS & Biomarker Result & Treatment & Expected treatment response & Reference \\
		\midrule
		\multirowcell{2}[4pt][l]{HER2} &
		\multirowcell{2}[0pt][l]{Expressed in 1/3 of\\UCS\supercite{Yoshida2023-bl,Jenkins2022-zi}} &
		\multirowcell{2}[0pt][l]{Positive\\(Histopathological\\report, January, 2021\\ "HER2 is positive in\\some of the tumor\\cells with aninter-\\mediate level of\\ positivity and an \\incomplete level \\of circumferential \\positivity.")} &
		Trastuzumab deruxtecan &
		The phase II STATICE trial enrolled 22 HER2-high and 10 HER2-low patients with recurrent UCS. Objective response rates (ORRs) were 54.5\% and 70\% in the HER2-high and HER2-low groups, respectively. Median progression-free survival (PFS) was 6.2 months for HER2-high patients and 13.3 months for HER2-low patients. Overall survival (OS) was 6.7 months for HER2-high patients and not reached for HER2-low patients. Three patients in each group had received at least three prior lines of therapy. &
		Phase II study \supercite{Nishikawa2023-hj} \\
		&
		&
		&
		T-DM1 &
		T-DM1 demonstrated significant antitumor activity in HER2-overexpressing CS xenograft models, resulting in prolonged survival compared to trastuzumab. &
		Preclinical evidence \supercite{Nicoletti2015-zg} \\
		\multirow{2}{*}{ER} &
		Estrogen receptor: 20–30\%; progestin receptor: 5–40\% \supercite{Bogani2023-qf} &
		80\% &
		Anithormonal treatment (e.g., anastrozole) &
		The phase II PARAGON study enrolled seven patients with UCS and evaluated anastrozole treatment. A clinical benefit rate (CBR) of 43\% was observed at three months, with a median duration of clinical benefit of 5.6 months. While stable disease was noted in three patients, no objective responses were achieved. Median progression-free survival was 2.7 months. 43\% of the entire cohort (UCS and leiomyosarcoma) had received prior chemotherapy. &
		Phase II study \supercite{Edmondson2021-co} \\
		&
		\textit{ESR1} Amplification &
		7\% &
		Pembrolizumab + lenvatinib + letrozole &
		A patient with metastatic, pMMR, and \textit{ESR1}-amplified UCS achieved a durable partial response of 36 months with third-line treatment combining pembrolizumab, lenvatinib, and letrozole. &
		Case report \supercite{Soiffer2024-dl} \\
		FR-alpha &
		Expressed in 1/3 of UCS \supercite{Saito2023-mh} &
		Not determined. &
		Mirvetuximab soravtansine + pembrolizumab &
		Patients with FR-alpha positive tumors may be eligible for combination therapy with mirvetuximab soravtansine and pembrolizumab. This approach is supported by results from a phase II trial (NCT03835819) demonstrating efficacy in pMMR-positive patients, including those with prior pembrolizumab treatment failure. Interim analysis in endometrial cancers showed ORR of 37.5\%. No stratified analysis for UCS available. &
		Phase II study \supercite{Porter2024-ck} \\
		HRD &
		Unknown &
		Not determined. &
		Poly (ADP-Ribose) Polymerase Inhibitor (PARPi) &
		UCS cell lines exhibiting HRD signature demonstrated significantly increased sensitivity to olaparib compared to homologous recombination proficient UCS cell lines, both in vitro and in vivo. &
		Preclinical evidence \supercite{Tymon-Rosario2022-me} \\
		MAGE-A4 &
		Expressed in 91\% of carcinosarcomas \supercite{Resnick2002-mj} &
		Not determined. &
		Bispecific T Cell Engaging Receptor Molecule targeting MAGE-A4/8 expression &
		Patients with MAGE-A4-positive UCS may be eligible for participation in an ongoing clinical trial located in Bavaria: A Phase Ia/Ib First-In-Human Clinical Trial to Evaluate the Safety, Tolerability and Initial Anti-tumor Activity of IMA401, a Bispecific T Cell Engaging Receptor Molecule (TCER®), in Patients With Recurrent and/or Refractory Solid Tumors. &
		\makecell[tl]{https://clinicaltrials.gov/\\study/NCT05359445} \\
		\multirow{2}{*}{PRAME} &
		\multirowcell{2}[-5pt][tl]{Expressed in 60\% of\\ UCS\supercite{Alrohaibani2024-ia,Roszik2017-rq}} &
		\multirow{2}{*}{Not determined.} &
		Bispecific T Cell-Engaging Receptor Molecule targeting PRAME &
		Patients with PRAME-positive UCS may be eligible for participation in two ongoing clinical trials located in Bavaria: a) IMA402-101: A Phase I/II First-In-Human Clinical Trial to Evaluate the Safety, Tolerability and Anti-Tumor Activity of IMA402, a Bispecific T Cell-Engaging Receptor Molecule (TCER) Targeting PRAME, in Patients With Recurrent and/or Refractory Solid Tumors. &
		\makecell[tl]{a) https://clinicaltrials.gov/\\study/NCT05958121 }\\
		&
		&
		&
		Genetically Modified Autologous T Cells Expressing a T Cell Engaging Receptor Recognizing PRAME as Monotherapy or in Combination with Nivolumab &
		b) IMA203-101: Phase 1 Study Evaluating Genetically Modified Autologous T Cells Expressing a TCR Recognizing a Cancer/Germline Antigen as Monotherapy or in Combination With Nivolumab in Patients With Recurrent and/or Refractory Solid Tumors &
		\makecell[tl]{b) https://clinicaltrials.gov/\\study/NCT03686124 }\\
		\multirow{2}{*}{Trop2} &
		\multirowcell{2}[0pt][l]{Expressed in 1/3 of\\UCS\supercite{Raji2011-ur,Lopez2020-pc}} &
		\multirow{2}{*}{Not determined.} &
		\multirow{2}{*}{Sacituzumab govitecan} &
		Twenty-two patients with Trop2-positive recurrent endometrial cancer were enrolled in a phase II study evaluating sacituzumab govitecan. Of these, three patients had UCS. Among the 20 response-evaluable patients, an objective response rate of 35\% was observed. Median progression-free survival (PFS) and overall survival (OS) were 5.7 months and 22.5 months, respectively. The median (range) number of prior anticancer regimens was 3 (1–6). &
		Phase II study \supercite{Santin2023-gm} \\
		&
		&
		&
		&
		Sacituzumab govitecan demonstrated significant tumor growth inhibition and improved 90-day overall survival in Trop2-positive carcinosarcoma cell lines compared to Trop2-negative controls. &
		Preclinical evidence \supercite{Lopez2020-pc} \\
		Serum CA-125 &
		Elevated serum CA-125 levels in 62\% of UCS patients with FIGO stage III or IV \supercite{Huang2007-aj} &
		Elevated at progression in 2021, normalized since 2022 correlating with partial response. &
		Treatment monitoring with Serum CA-125 &
		Preoperative CA-125 elevation correlates with extrauterine disease and deep myometrial invasion in patients with UCS. Postoperatively, elevated CA-125 is an independent prognostic indicator of poor survival. These findings suggest that CA-125 could serve as a valuable serum marker for managing UCS patients. &
		Retrospective analysis \supercite{Huang2007-aj} \\
		\bottomrule
	\end{tabular}
	} 
	\label{tab:table3}
\end{table}

\printbibliography

\newpage
\section*{Supplements}

\begin{supfigure}[H]
	\includegraphics[width=\linewidth]{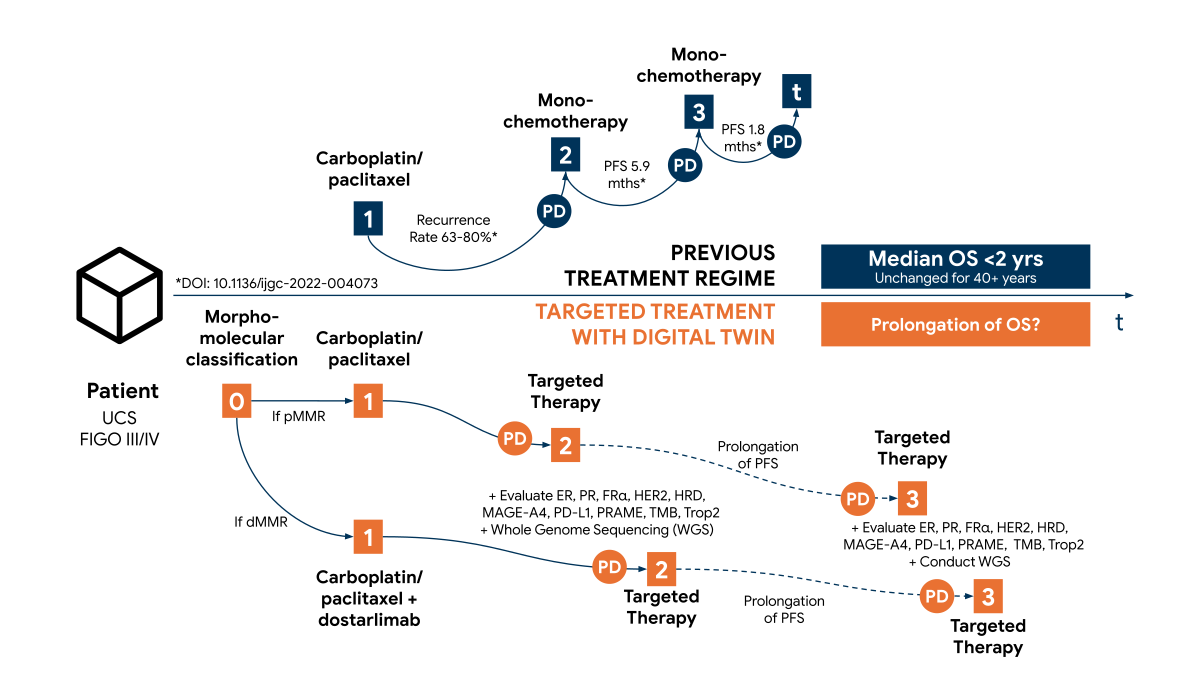}
	\caption{The previous treatment regime did not mention the addition of dostarlimab as the consensus statement was published before EMA approval of dostarlimab on October 12, 2023. Reported response rates and median PFS with third-line therapy were 5.5\% and 1.8 months, respectively. The 5-year overall survival rate has not changed in decades (31.9\% in 1975 to 33.8\% in 2012) \supercite{Bogani2023-qf}. The targeted treatment approach could result in a prolongation of PFS and consequently a better OS.}
	\label{fig:supfig1}
	
\end{supfigure}
\newpage
\begin{suptable}[H]
	\caption{Detailed results of MTB for case 1}
	\centering
	\scalebox{.6}{
		\begin{tabular}{|p{3.5cm}|p{3.5cm}|p{2.5cm}|p{2.5cm}|p{2.5cm}|p{3cm}|p{3cm}|p{3cm}|}
			\toprule
			Submitting hospital &
			Block material &
			Biomarkers tested before MTB &
			Tumor cell count &
			Panel used &
			Single Nucleotide Variants (SNV)/Indel &
			Copy number changes &
			Splice variations/ translocations \\
			\midrule
			TUM University Hospital, Munich, Bavaria, Germany &
			Supracervical lymph node metastasis (2021) &
			ER: 80\% PR: 3\% HER2: Positive &
			60\% &
			TSO 500 (DNA)/ TST 170 (RNA) &
			None detected &
			None detected &
			None detected \\
			\midrule
			Gene             & Reference number & Exon & cDNA                   & Protein     & Allele frequency & COSMIC database v90 & Class \\
			\midrule
			\textit{TP53}    & NM\_000546       & 4    & c.150delT              & p.I50Mfs*73 & 14\%             & COSV52758078        & 4     \\
			\textit{PPP2R1A} & NM\_014225       & 6    & c.771G\textgreater{}T  & p.W257C     & 12\%             & COSV59043009        & 4     \\
			\textit{NOTCH4}  & NM\_004557       & 18   & c.2780G\textgreater{}A & p.C927Y     & 8\%              & No entry            & 3     \\
			\textit{RUNX1}   & NM\_001754       & 9    & c.1070C\textgreater{}T & p.P357L     & 9\%              & No entry            & 3     \\
			\textit{AR}      & NM\_000044       & 1    & c.476C\textgreater{}G  & p.A159G     & 7\%              & No entry            & 3    \\
			\bottomrule
		\end{tabular}
	}
	\label{tab:suptable1}
\end{suptable}

\begin{suptable}[H]
	\caption{Data structure for RGT Digital Twin}
	\centering
	\scalebox{.8}{
		\begin{tabular}{|l|l|p{8cm}|}
			\toprule
			Attribute           & Data Type & Description                                                                                                   \\
			\midrule
			n                   & Integer   & Number of patients in the study                                                                               \\
			age                 & String    & Age of patients in the study; string as many studies contain ranges                                           \\
			gender              & String    & Gender(s) of patients in the study                                                                            \\
			race                & String    & Race(s) of patients in the study                                                                              \\
			diagnosis           & String    & Diagnosis of patients in the study                                                                            \\
			biomarkers &
			\begin{tabular}[t]{@{}l@{}}\{\\  'pd-l1':  String,\\  'tmb/mb': String,\\  'msi/mss':String,\\  'others': String\\ \}\end{tabular} &
			Biomarkers (e.g., PD-L1) determined and discussed in the study; returned as dictionary for simplified analysis \\
			previous treatments & String    & Description of previous treatments (and response)                                                             \\
			study treatment     & String    & Treatments discussed in the study                                                                             \\
			study treatment response &
			\begin{tabular}[t]{@{}l@{}}\{\\  'treatment response': String,\\  'adverse effects':    String\\ \}\end{tabular} &
			Response to treatments discussed in the study including adverse effects; returned as dictionary for simplified analysis \\
			PFS                 & String    & Progression-free survival (PFS) in months reported for study treatment; string as many studies contain ranges \\
			OS                  & String    & Overall survival (OS) in months reported for patient cases; string as many studies contain ranges \\  
			\bottomrule
		\end{tabular}
	}
	\label{tab:suptable2}
\end{suptable}
\begin{landscape}
	\begin{suptable}[H]
		\caption{Baseline characteristics of institutional MTB cases at TUM University Hospital}
		\centering
		\scalebox{.6}{
			\begin{tabular}{|l|l|p{4cm}|p{1.3cm}|p{4cm}|p{4cm}|l|p{2cm}|l|l|l|l|}
				\toprule
				ID &
				Gender &
				Primary site &
				Age at diagnosis &
				Stage at diagnosis &
				Sites of metastases &
				Time of MTB &
				Site sequenced &
				Tumor cell count &
				PD-L1 &
				TMB (Mut/Mb) &
				MMR \\
				\midrule
				1 &
				female &
				Uterine Carcinosarcoma (UCS) &
				66 &
				FIGO IIIC2, pT3a, pN2 (3/67), R0 &
				Cervical lymph node metastasis, pelvic recurrence, retroperitoneal lymph node metastases &
				2021 &
				Cervical lymph node metastasis &
				60\% &
				CPS: 41 TPS: 3\% IC: 40\% &
				6.3 (intermediate) &
				pMMR/MSS (3.6\%) \\
				2 &
				female &
				Cervical Squamous Cell Caricnoma (CESC) &
				30 &
				FIGO IB1, pT1b1, pNX, G3, R0 &
				Pelvic recurrence, liver metastasis &
				2020 &
				Primary surgery &
				70\% &
				CPS: 75 TPS: 70\% IC: 5\% &
				0 (low) &
				pMMR/MSS (1.11\%) \\
				3 &
				female &
				Cervical Squamous Cell Caricnoma (CESC) &
				31 &
				FIGO IVB, cT2a, pN1 (12/94), cM1 (PER) &
				Peritoneal metastases &
				2020 &
				Peritoneal metastasis &
				75\% &
				CPS: 40 TPS: 40\% IC: \textless{}1\% &
				3.1 (low) &
				pMMR/MSS (0\%) \\
				4 &
				female &
				Cervical Squamous Cell Caricnoma (CESC) &
				79 &
				FIGO IVa, cT2b2, cN1, cM0, G2 &
				Lymph node metastases &
				2021 &
				Primary tumor biopsy &
				80\% &
				CPS: 81 TPS: 80\% IC: 1\% &
				11 (intermediate) &
				pMMR/MSS (4.6\%) \\
				5 &
				female &
				Cervical Adenocarcinoma (CEAD) &
				32 &
				FIGO IVB, cT2b, cN1, cM1 (LYM) &
				Lymph node metastases &
				2021 &
				Ileocecal resection &
				40\% &
				CPS: 95 TPS: 90\% IC: 5\% &
				5.5 (intermediate) &
				pMMR/MSS (3.28\%) \\
				6 &
				female &
				Uterine Serous Carcinoma (USC) &
				55 &
				FIGO IA1, pT1a, pNx, L0, V0, Pn0, R0 &
				Peritoneal metastases, lymph node metastases &
				2024 &
				Inguinal lymph node metastasis &
				30\% &
				CPS: 40 TPS: 30\% IC: 8\% &
				13.4 (intermediate) &
				pMMR/MSS (1.89\%) \\
				7 &
				male &
				Undifferentiated Sarcomatoid Carcinoma of the Pancreas &
				59 &
				pT3, pN1 (3/81), L1, V1, Pn1, R0 &
				Lcooregional recurrence, liver metastases &
				2022 &
				Liver metastasis &
				70\% &
				CPS: 85 TPS: 80\% IC: 4\% &
				3.2 (low) &
				pMMR/MSS (2.61\%)\\ 
				\bottomrule
			\end{tabular}
		}
		\label{tab:suptable3}
	\end{suptable}
	
	\begin{suptable}[H]
		\caption{ Seven studies on ICI treatment in UCS lacked patient-level data necessary for individual digital twin creation}
		\centering
		\scalebox{.6}{
			\begin{tabular}{|p{2.5cm}|p{2.5cm}|p{2.5cm}|p{4cm}|p{4cm}|p{4cm}|p{5cm}|p{4cm}|p{4cm}|}
				\toprule
				Trial &
				Recruitment period &
				Phase &
				Experimental group &
				Control group &
				Sample size &
				Treatment response &
				Median follow-up (months) &
				Median PFS (months) experimental vs control \\
				\midrule
				RUBY trial \supercite{Mirza_Mansoor_R2023-jg} &
				2019-2021 &
				III &
				Carboplatin/paclitaxel (CP) + dostarlimab x 6 cycles + maintenance with dostarlimab &
				CP + placebo x 6 cycles + maintenance with placebo &
				\begin{tabular}[t]{@{}l@{}}Overall: 494UCS: n = 44\\ \\ dMMR: 118(UCS: 5)\\pMMR: 376(UCS: 39)\end{tabular} &
				\begin{tabular}[t]{@{}l@{}}Not stratified for UCS\\ Overall: Hazard Ratio\\ (HR) = 0.64 (0.51-0.80) \\p \textless{}0.001\\ dMMR: HR = 0.28 (0.16-0.50) \\p \textless{}0.001\\ pMMR: HR = 0.76 (0.59-0.98)\end{tabular} &
				\begin{tabular}[t]{@{}l@{}}Not stratified for UCS\\ Overall: 25.4\\ dMMR: 24.8\\ pMMR NA\end{tabular} &
				NA \\
				DUO-E trial \supercite{Westin2024-yq} &
				2020-2022 &
				III &
				CP + durvalumab x 6 cycles + maintenance with durvalumab &
				CP + placebo x 6 cycles + maintenance with placebo &
				\begin{tabular}[t]{@{}l@{}}Overall 479\\ UCS: n = 61\\ dMMR 95\\ pMMR 384\end{tabular} &
				\begin{tabular}[t]{@{}l@{}}Overall: HR = 0.71 (0.57-0.89) \\p = 0.003\\ dMMR: HR = 0.42 (0.22-0.80)\\ pMMR: HR = 0.77 (0.60-0.97)\\ Histology: other, \\including UCS (27/39)\\ HR = 0.76 (0.46-1.25), n.s.\end{tabular} &
				\begin{tabular}[t]{@{}l@{}}Not stratified for UCS\\ Control 12.6\\ Experimental 15.4\\ dMMR 10.2\\ pMMR 12.8\end{tabular} &
				\begin{tabular}[t]{@{}l@{}}Not stratified for UCS\\ Overall 10.2 vs 9.6\\ dMMR NR vs 7\\ pMMR 9.9 vs. 9.7\end{tabular} \\
				AtTEnd trial \supercite{Colombo2024-tp} &
				2018-2022 &
				III &
				CP + atezolizumab x 6 cycles + maintenance with atezolizumab &
				CP + placebo x 6 cycles + maintenance with placebo &
				\begin{tabular}[t]{@{}l@{}}Overall: 549\\ UCS: n = 50\\ dMMR: 125\\ pMMR: 409\end{tabular} &
				\begin{tabular}[t]{@{}l@{}}Overall: HR = 0.74 (0.61-0.91),\\p = 0.02\\ UCS: HR = 0.88 (0.45-1.73), n.s.\\ \\ dMMR: HR = 0.36 (0.23-0.57), \\p = 0.0005\\ UCS: HR = 0.41 (0.03-6.62), n.s.\\ pMMR: HR = 0.92 (0.73-1.16), n.s.\\ UCS: not specified\end{tabular} &
				\begin{tabular}[t]{@{}l@{}}Not stratified for UCS\\ Overall: 28.3\\ dMMR: 26.2\\ pMMR NA\end{tabular} &
				\begin{tabular}[t]{@{}l@{}}Not stratified for UCS\\ Overall: 10.1 vs 8.9\\ dMMR NR vs. 6.9\\ pMMR 9.5 vs 9.2\end{tabular} \\
				Single-center, randomized, open-label, phase II trial 33 &
				\begin{tabular}[t]{@{}l@{}}N/a\\ Data cut-off:\\December 2021\end{tabular} &
				II &
				\begin{tabular}[t]{@{}l@{}}Arm 1: Durvalumab\\ Arm 2: Durvalumab +\\ tremelimumab\end{tabular} &
				None &
				\begin{tabular}[t]{@{}l@{}}Overall: 82\\ UCS: 16\\ Arm 1: 6\\ Arm 2: 10\end{tabular} &
				\begin{tabular}[t]{@{}l@{}}Overall:\\ Arm 1 Overall Response Rate\\(ORR): 10.8\%\\ Arm 2 ORR: 5.3\%\\ UCS: ORR: 0\%\end{tabular} &
				N/a &
				N/a \\
				NCI-MATCH (EAY131) \supercite{Azad2020-bz} &
				2016-2017 &
				II &
				Nivolumab &
				None &
				\begin{tabular}[t]{@{}l@{}}Overall: 42\\ UCS: n = 4\end{tabular} &
				\begin{tabular}[t]{@{}l@{}}Not stratified for UCS\\ Overall ORR: 36\%\end{tabular} &
				17.3 &
				\begin{tabular}[t]{@{}l@{}}Not stratified for UCS\\ 6-month PFS rate: 51.3\%\\ 12-month PFS rate: 46.2\%\\ 18-month PFS rate: 31.4\%\end{tabular} \\
				Retrospective institutional analysis from The University of Texas MD Anderson Cancer Center \supercite{How2021-tl} &
				2019-2020 &
				Retrospective study of institutional data &
				\begin{tabular}[t]{@{}l@{}}Pembrolizumab + lenvatinib\\ Recommended dose of\\lenvatinib vs. reduced\\dose of lenvatinib\end{tabular} &
				None &
				\begin{tabular}[t]{@{}l@{}}Overall: n = 61\\ Recommended dose: n = 14\\ Reduced dose: n = 47\\ UCS: n = 16\\ Recommended dose: n = 3\\ Reduced dose: n = 13\end{tabular} &
				\begin{tabular}[t]{@{}l@{}}ORR:\\ Overall: 36.1\%\\ UCS: 25\% (3/12)\\ Clinical benefit rate (CBR):\\ Overall: 68.9\%\\ UCS: 58.3\% (7/12)\end{tabular} &
				\begin{tabular}[t]{@{}l@{}}Not stratified for UCS\\ Overall:\\ Recommended dose: 3.2\\ Reduced dose: 5.5\end{tabular} &
				\begin{tabular}[t]{@{}l@{}}Not stratified for UCS\\ Recommended dose: 8.6\\ Reduced dose: 9.4\end{tabular} \\
				Multicenter, randomized, phase II trial \supercite{Lheureux2022-bf} &
				2018-2019 &
				II &
				Cabozantinib + nivolumab &
				None &
				\begin{tabular}[t]{@{}l@{}}Arm A: 36\\ Arm B: 18\\ Arm C:\\ UCS: n = 10\\ pMMR: 100\%\end{tabular} &
				\begin{tabular}[t]{@{}l@{}}Arm C (UCS):\\ ORR: 10\%\\ 1 PR, 5 SD\end{tabular} &
				Overall (Arm A, B, C): 15.9 &
				\begin{tabular}[t]{@{}l@{}}Arm C (UCS):\\ Median SD duration:\\3.2 (range 2.8-7.6)\end{tabular}\\
				\bottomrule
			\end{tabular}
			\label{tab:suptable4}
		} 
	\end{suptable}
\end{landscape}

\end{document}